\documentclass[letterpaper, 10 pt, conference]{ieeeconf}  

\IEEEoverridecommandlockouts                              

\usepackage{amsmath}
\usepackage{url}
\usepackage{caption}
\usepackage{subcaption}
\usepackage{graphicx}
\usepackage{multirow}
\usepackage{etoolbox}
\usepackage{placeins}
\usepackage{hyperref}
\usepackage{setspace} 
\usepackage{bbding}
\usepackage{xcolor}
\usepackage{bbold}
\usepackage[normalem]{ulem}
\useunder{\uline}{\ul}{}
\usepackage{nicematrix}
\usepackage[outline]{contour}
\newcommand{\cmark}{\ding{51}}%
\newcommand{\xmark}{\ding{55}}%
\DeclareMathAlphabet\mathbfcal{OMS}{cmsy}{b}{n}
\usepackage{arydshln}
\usepackage{tikz}
\usepackage{balance}

\usepackage{amssymb}
\usepackage{wasysym}
\usepackage{pifont}

\definecolor{Green}{RGB}{0, 255, 0}
\definecolor{Red}{RGB}{255, 0, 0}
\definecolor{myblue}{RGB}{0, 0, 255}
\definecolor{myyellow}{RGB}{255, 255, 0}
\definecolor{mypurple}{RGB}{68, 1, 84}

\newcommand{\greensquare}{\raisebox{0pt}{\tikz{\fill[Green] (-0.1,-0.1) rectangle (0.1,0.1);}}}
\newcommand{\redsquare}{\raisebox{0pt}{\tikz{\fill[Red] (-0.1,-0.1) rectangle (0.1,0.1);}}}
\newcommand{\bluesquare}{\raisebox{0pt}{\tikz{\fill[myblue] (-0.1,-0.1) rectangle (0.1,0.1);}}}
\newcommand{\transparentdotted}{\raisebox{0pt}{\tikz{\draw[black, densely dotted] (-0.1,-0.1) rectangle (0.1,0.1);}}}
\newcommand{\yellowsquare}{\raisebox{0pt}{\tikz{\fill[myyellow] (-0.1,-0.1) rectangle (0.1,0.1);}}}
\newcommand{\purplesquare}{\raisebox{0pt}{\tikz{\fill[mypurple] (-0.1,-0.1) rectangle (0.1,0.1);}}}

\title{\LARGE \bf
High-Quality Unknown Object Instance Segmentation via \\Quadruple Boundary Error Refinement
}

\author{
Seunghyeok Back$^{1}$,\hspace{-3pt} Sangbeom Lee$^{2}$,\hspace{-3pt} Kangmin Kim$^{2}$,\hspace{-3pt} Joosoon Lee$^{2}$,\hspace{-3pt} Sungho Shin$^{3}$,\hspace{-3pt} Jemo Maeng$^{2}$,\hspace{-3pt} Kyoobin Lee$^{2\dagger}$%
\thanks{$^{1}$ S. Back is with the Department of AI Machinery, Korea Institute of Machinery \& Materials (KIMM), Daejeon 34103, Republic of Korea.}  
\thanks{$^{2}$ S. Lee, K. Kim, J. Lee, J. Maeng, and K. Lee are with the Department of AI Convergence, Gwangju Institute of Science and Technology (GIST), Gwangju 61005, Republic of Korea.}  
\thanks{$^{3}$ S. Shin is with the Robotics Lab, Hyundai Motor Company, Uiwang 16082, Republic of Korea.} 
\thanks{S. Back and S. Shin were with GIST at the time of the initial submission.}  
\thanks{$^{\dagger}$ Corresponding author: Kyoobin Lee {\tt\small kyoobinlee@gist.ac.kr}}  
}

\begin{document}

\maketitle
\thispagestyle{empty}
\pagestyle{empty}

\begin{abstract}

Accurate and efficient segmentation of unknown objects in unstructured environments is essential for robotic manipulation. Unknown Object Instance Segmentation (UOIS), which aims to identify all objects in unknown categories and backgrounds, has become a key capability for various robotic tasks. However, existing methods struggle with over-segmentation and under-segmentation, leading to failures in manipulation tasks such as grasping. To address these challenges, we propose QuBER (Quadruple Boundary Error Refinement), a novel error-informed refinement approach for high-quality UOIS. QuBER first estimates quadruple boundary errors—true positive, true negative, false positive, and false negative pixels—at the instance boundaries of the initial segmentation. It then refines the segmentation using an error-guided fusion mechanism, effectively correcting both fine-grained and instance-level segmentation errors. Extensive evaluations on three public benchmarks demonstrate that QuBER outperforms state-of-the-art methods and consistently improves various UOIS methods while maintaining a fast inference time of less than 0.1 seconds. Furthermore, we show that QuBER improves the success rate of grasping target objects in cluttered environments. Code and supplementary materials are available at \url{https://sites.google.com/view/uois-quber}.
\end{abstract}


\section{Introduction}

The ability to perceive and manipulate unknown objects in cluttered environments is crucial for robotics and embodied AI. At the core of this challenge is Unknown Object Instance Segmentation (UOIS) \cite{danielczuk2019segmenting, xie2020best, xie2021unseen}, which enables robots to identify and interact with novel objects in unstructured settings. UOIS aims to detect all object instances in unknown categories and backgrounds and has become fundamental to various robotic manipulation tasks, including grasping, pushing, and rearrangement \cite{yu2022self, sundermeyer2021contact, goyal2022ifor}. The quality of UOIS directly impacts task success, as inaccurate segmentation often leads to failures in subsequent robotic operations.

\begin{figure}[!ht]
\centering
     \includegraphics[width=1.0\columnwidth]{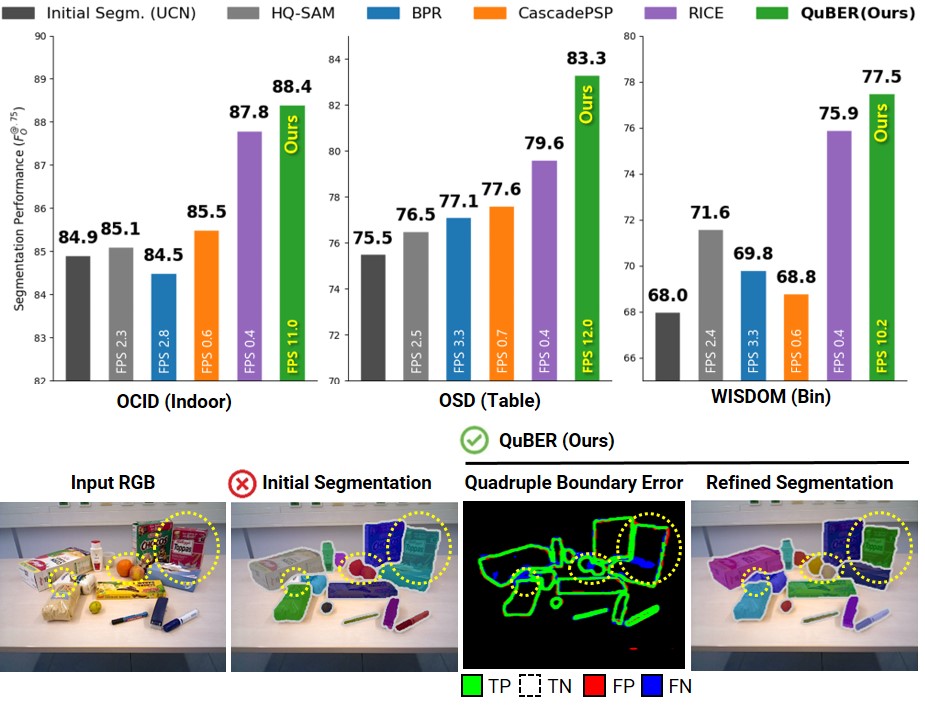}
\caption{(Top) Results of the proposed QuBER method for high-quality UOIS across various domains. (Bottom) From the initial segmentation, QuBER performs error-informed refinement by estimating pixel-wise quadruple boundary errors and refining the segmentation based on these error estimates.}
\label{fig1}
\end{figure}

State-of-the-art UOIS methods \cite{danielczuk2019segmenting, xie2020best, xiang2021learning, xie2021unseen, back2022unseen, lu2024mean} directly predict segmentation by leveraging both texture and geometric cues from RGB-D images. Although promising, these methods often struggle in complex, cluttered scenes with significant occlusions, leading to instance-level errors such as over-segmentation (splitting a single object) or under-segmentation (merging multiple objects). Segmentation refinement methods \cite{lin2017refinenet, yuan2020segfix, cheng2020cascadepsp, tang2021look, xie2022rice} improve local details in boundaries but typically assume instance-level correctness, limiting their ability to rectify major errors. Recent promptable networks, such as Segment Anything Model (SAM) \cite{kirillov2023segment, ke2024segment, ren2024grounded}, trained on web-scale datasets to achieve strong generalizability, show promise for both initial segmentation and refinement. However, in real-world robotic cluttered scenes, they often over-segment \cite{zhang2024improving, fang2024embodied} or refine only minor details without accurate human prompts. Additionally, these methods are computationally intensive, making them less suitable for real-time robotic applications.

In this paper, we propose \textbf{Qu}adruple \textbf{B}oundary \textbf{E}rror \textbf{R}efinement (\textbf{QuBER}), a novel model for high-quality UOIS using an error-informed refinement strategy (Fig. \ref{fig1}). QuBER estimates quadruple boundary errors - true positive (TP), true negative (TN), false positive (FP), and false negative (FN) pixels at instance boundaries. Our Error Guidance Fusion (EGF) module then incorporates these estimated errors to refine the segmentation accurately. Notably, quadruple boundary error estimation effectively captures both fine-grained and instance-level errors, providing explicit error correction information (e.g., FP indicates over-segmented regions to be deleted, FN indicates under-segmented regions to be added). Unlike existing methods that focus solely on local refinements or require computationally expensive operations, our error-informed approach efficiently corrects segmentation errors with a fast inference time ($\sim$0.1s on RTX3090, Intel6248R). Extensive experiments demonstrate that QuBER outperforms state-of-the-art UOIS and refinement methods on three benchmarks by resolving over-segmentation and under-segmentation issues across various initial segmentation methods. We also demonstrate the practical robot application of QuBER by improving target object grasping success.

The main contributions of this study are as follows: (1) We introduce QuBER, an error-informed refinement network for high-quality UOIS. (2) We propose quadruple boundary error estimation to capture and refine both fine-grained and instance-level errors. (3) We propose an Error Guidance Fusion (EGF) module for effectively integrating estimated errors into the refinement process.

\section{Related Work}

\subsection{Unknown Object Instance Segmentation}

Unknown Object Instance Segmentation (UOIS) \cite{danielczuk2019segmenting, xie2020best, xie2021unseen} aims to detect all arbitrary object instances in images with unknown objects and environments, serving as a fundamental perception module for robotic manipulation \cite{yu2022self, sundermeyer2021contact, goyal2022ifor}. Early approaches relied on clustering techniques \cite{felzenszwalb2004efficient, richtsfeld2012segmentation}, while recent state-of-the-art methods employ deep networks trained on large-scale synthetic RGB-D data to learn category-agnostic objectness \cite{danielczuk2019segmenting, xie2020best, back2020segmenting, xiang2021learning, xie2021unseen, back2022unseen, lu2024mean}. Prompt-based segmentation models like SAM, trained on web-scale data \cite{kirillov2023segment, ke2024segment, ren2024grounded}, show promise but suffer from over-segmentation without accurate prompts \cite{zhang2024improving, fang2024embodied} and incur high computational costs during inference. Despite these advancements, high-quality object segmentation in cluttered environments remains challenging due to common issues like over-segmentation and under-segmentation \cite{xie2020best, xiang2021learning, back2022unseen, lu2024mean}. Our method addresses these issues by refining existing UOIS outputs, thereby enhancing segmentation quality for robust manipulation in complex scenes.

\subsection{Refining Segmentation}

Object segmentation quality directly impacts robotic manipulation tasks like grasping \cite{sundermeyer2021contact}. Conditional random fields \cite{krahenbuhl2011efficient, chen2017deeplab} refine segmentation by modeling spatial relationships but struggle with semantic information and large error regions. Recent methods \cite{kirillov2020pointrend, shen2021dct} focus on fine-grained improvements: Segfix \cite{yuan2020segfix} replaces unreliable boundary predictions with inner predictions, Boundary Patch Refinement (BPR) \cite{tang2021look} predicts boundary-aligned masks from patched coarse masks, and CascadePSP \cite{cheng2020cascadepsp} adopts a cascade strategy for pixel-aligned refinement. However, these methods primarily address fine-grained boundary details without resolving instance-level over-segmentation and under-segmentation errors. SAM \cite{kirillov2023segment}-based methods such as HQ-SAM \cite{ke2024segment} show promise but are computationally expensive and focus mainly on fine details. RICE \cite{xie2022rice} tackles instance-level refinement through perturbation and sampling but requires about 4 seconds per frame.
Test-time adaptation \cite{zhang2023unseen} improves domain generalization but needs 20 seconds for new scenarios. In contrast, our approach achieves state-of-the-art performance with fast refinement in less than 0.1 seconds per frame, efficiently addressing fine-grained and instance-level errors.

\subsection{Error Detection for Segmentation and Refinement} 
Estimating errors in deep segmentation models is crucial for system reliability \cite{rahman2021run}. Most approaches focus solely on detecting binary errors (true, false), employing techniques such as maximum softmax probability \cite{hendrycks2016baseline}, Monte Carlo dropout \cite{gal2016dropout}, or error segmentation \cite{rahman2022fsnet, sun2022see}. Only a few works leverage estimated errors for segmentation refinement. SESV \cite{xie2020sesv} proposed a four-step framework (segmentation, evaluation, refinement, and verification), ERA \cite{kuhn2022reverse} introduced an error-reversing autoencoder, and failure detection and label correction networks were employed in \cite{ali2023learning}. However, these methods require secondary networks for error detection and refinement, which incur substantial computational costs, and focus on refining fine-grained details by predicting binary errors in semantic segmentation. In contrast, our approach introduces an error-informed refinement method for instance segmentation that unifies error estimation and refinement within a single network. Additionally, we propose quadruple boundary error estimation (TP, TN, FP, FN) for both fine-grained and instance-level error correction.

\section{Quadruple Boundary Error Refinement}

In this paper, we propose QuBER (Fig. \ref{fig3}), an error-informed refinement network for high-quality UOIS that addresses fast and accurate refinement of over-segmentation and under-segmentation in complex, cluttered scenes. 

\subsection{Error-informed Refinement}

Our goal is to design a segmentation refinement model $\mathcal{G}: (I, M_{i}) \rightarrow M_{r}$ that produces refined, high-quality masks $M_{r}$ for unknown object instances from an RGB-D image $I$ and initial segmentation (IS) masks $M_{i}$ of arbitrary initial segmentation models. To achieve this, we introduce an error-informed refinement process comprising the following steps:


\begin{itemize}

\item  \textbf{IS feature extractor} $\mathcal{F}: (I, M_{i}) \rightarrow h$ to obtain initial segmentation features $h$ from the input RGB-D image $I$ and coarse masks $M_{i}$.

\item  \textbf{Error estimator} $\mathcal{E}: h \rightarrow \hat{e}_i$ to predict segmentation errors $\hat{e}_i$ in the initial segmentation,  providing critical error information to guide the refinement process.

\item  \textbf{Error-informed refiner} $\mathcal{H}: (h, \hat{e}_{i}) \rightarrow M_r$ to integrate predicted errors $\hat{e}_i$ with IS features to produce refined segmentation $M_r$, focusing on erroneous regions for accurate and targeted refinement.
\end{itemize}

The error-informed refinement process is formulated as:
\begin{equation}
\mathcal{G}(I, M_{i}) = \mathcal{H}(h, \mathcal{E}(h)) = M_r, \text{where } h=\mathcal{F}(I, M_i)
\end{equation} 
For both training and inference, these modules are integrated into a single QuBER network and trained jointly in an end-to-end manner.  This unified approach enhances segmentation by explicitly estimating errors in the initial results, enabling targeted refinements for more accurate corrections. By sharing the IS feature extractor, the architecture reduces computational overhead, ensuring fast and efficient refinement.

\subsection{Quadruple Boundary Error Estimation}

\begin{figure}[!t]
\centering
    \captionsetup[subfloat]{labelfont=scriptsize,textfont=scriptsize,justification=centering}
    \subfloat[RGB]{\includegraphics[width=0.24\columnwidth]{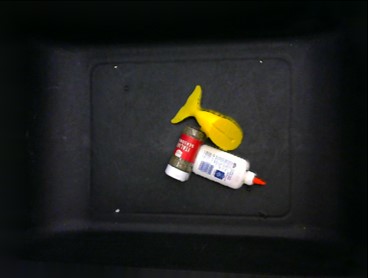}%
    \label{fig2_a}}
    \hspace{0.05mm}
    \vspace{-0.05mm}
    \subfloat[Initial. Segm.]{\includegraphics[width=0.24\columnwidth]{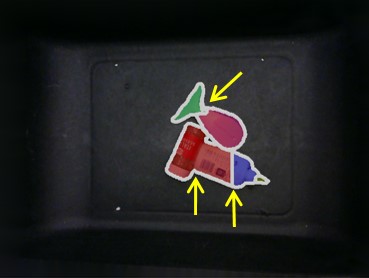}%
    \label{fig2_b}}
    \hspace{0.05mm}
    \subfloat[Refined Segm. (Ours)]{\includegraphics[width=0.24\columnwidth]{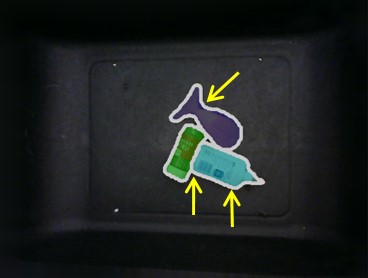}%
    \label{fig2_c}}
    \hspace{0.05mm}
    \subfloat[GT]{\includegraphics[width=0.24\columnwidth]{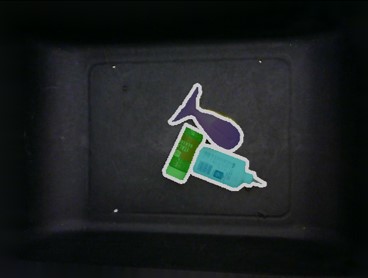}%
    \label{fig2_d}}
    \vspace{1mm}
    
    \subfloat[Binary Mask Error]{\includegraphics[width=0.32\columnwidth]{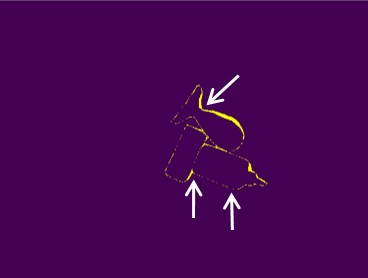}%
    \label{fig2_e}}
    \hspace{0.05mm}
    \subfloat[Quadruple \\Boundary Error (GT)]{\includegraphics[width=0.32\columnwidth]{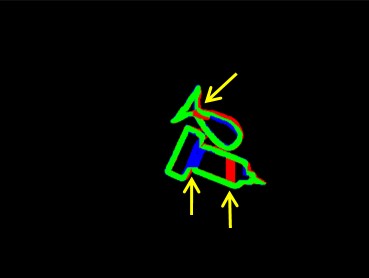}%
    \label{fig2_f}}
    \hspace{0.05mm}
    \subfloat[Quadruple \\Boundary Error (Pred.)]{\includegraphics[width=0.32\columnwidth]{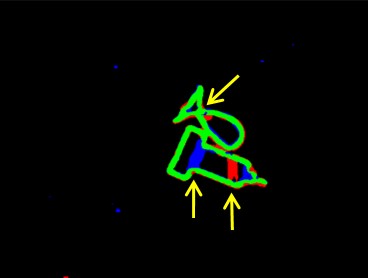}%
    \label{fig2_g}}
    
\caption{Comparison of binary mask errors (\protect\purplesquare{} True, \protect\yellowsquare{} False, shown in (e)) and our quadruple boundary errors (\protect\greensquare{} TP, \protect\transparentdotted{} TN, \protect\redsquare{} FP, \protect\bluesquare{} FN, shown in (f) and (g)) to represent segmentation errors in the initial segmentation (b) for error estimation. The proposed quadruple boundary error estimation effectively captures instance-level errors and facilitates precise refined segmentation (c).
}
\label{fig2}
\end{figure}

We employ quadruple boundary errors as the segmentation error $e_{i}$ in our error estimator (Fig. \ref{fig2}). Quadruple boundary errors are error pixel maps on instance boundaries with four categories: True Positive (TP), True Negative (TN), False Positive (FP), and False Negative (FN). This approach can effectively capture both fine-grained and instance-level errors while providing clear refinement guidance. The computation of quadruple boundary errors involves three steps: First, we obtain the boundaries of all IS and GT instance masks using dilation. Next, we create an instance boundary map by taking the union of these boundaries. Finally, we evaluate pixel-wise TP, TN, FP, and FN errors between the IS and GT boundary maps, resulting in an error map shape of $w\times h \times 4$, where $w$ and $h$ are the width and height, respectively.

This quadruple boundary error estimation (Fig. \ref{fig2_f}, and \ref{fig2_g}) offers significant advantages over standard binary mask errors (true, false). By focusing on boundary errors, we effectively target the most critical areas for refinement, as object boundaries are where most segmentation errors occur. Additionally, it captures instance-level errors such as under-segmentation and over-segmentation in overlapping instances. Moreover, it provides specific guidance for refinement by clearly indicating accurately segmented (TP, TN), over-segmented (FP), or under-segmented (FN) pixels, leading to effective refinement, as demonstrated in Fig. \ref{fig2_c}. In contrast, existing methods using binary mask errors (Fig. \ref{fig2_e}) struggle to capture instance-level errors and provide limited guidance for refinement, indicating only correct or incorrect pixels, which possibly leads to ambiguity in the refinement.

\label{BQE}

\begin{figure}[tp]
\smallskip
\smallskip
\centering
  \includegraphics[width=0.5\textwidth]{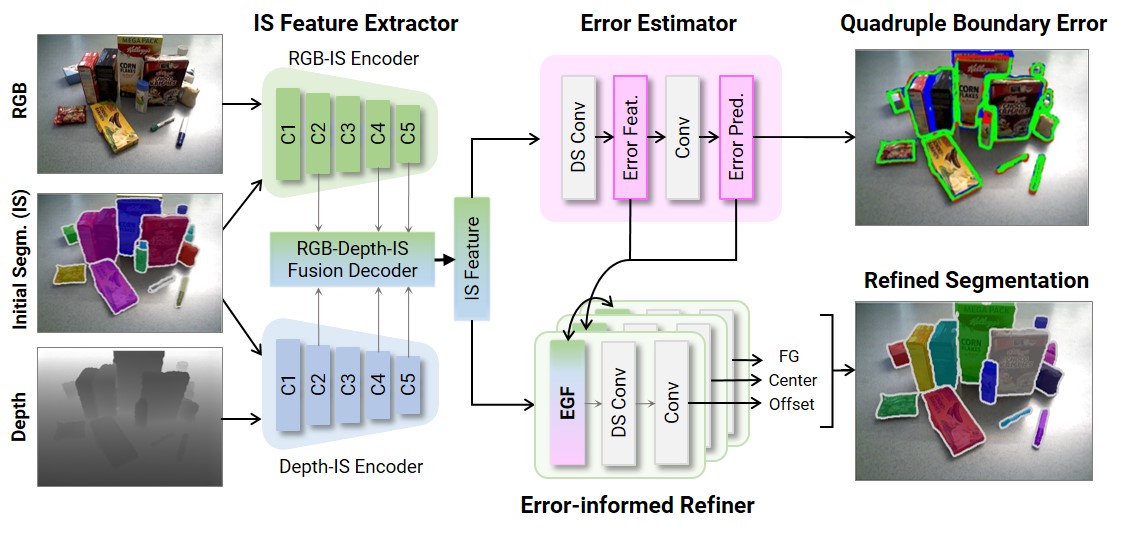}
  \caption{
   Overview of QuBER for error-informed refinement. 
  }
  \label{fig3}
\end{figure}
\begin{figure}[!t]
\centering
    \captionsetup[subfloat]{labelfont=tiny,textfont=tiny,justification=centering}
    \subfloat[RGB]{\includegraphics[width=0.155\columnwidth]{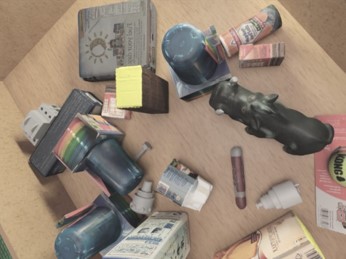}%
    \label{fig4_a}}
    \hspace{0.001mm}
    \subfloat[GT Masks]{\includegraphics[width=0.155\columnwidth]{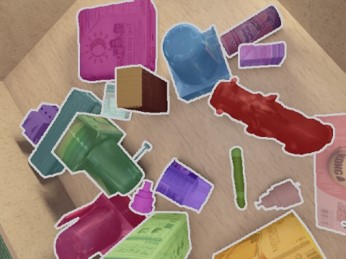}%
    \label{fig4_b}}
    \hspace{0.001mm}
    \subfloat[Perturbed Masks]{\includegraphics[width=0.155\columnwidth]{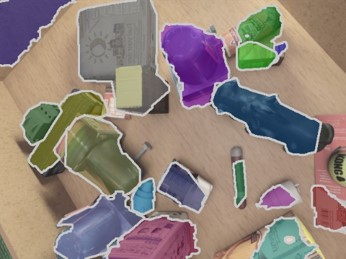}%
    \label{fig4_c}}
    \hspace{0.001mm}
    \subfloat[Center Map]{\includegraphics[width=0.155\columnwidth]{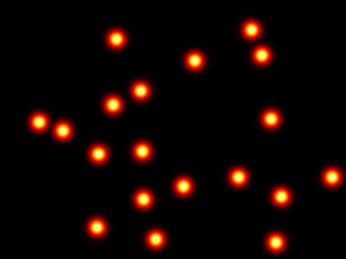}%
    \label{fig4_d}}
    \hspace{0.001mm}
    \subfloat[Offset Map]{\includegraphics[width=0.155\columnwidth]{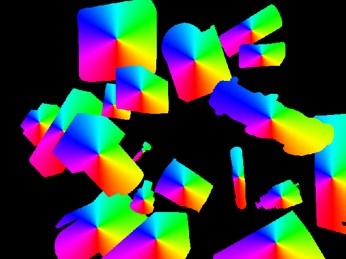}%
    \label{fig4_e}}
    \hspace{0.001mm}
    \subfloat[Foreground Mask]{\includegraphics[width=0.155\columnwidth]{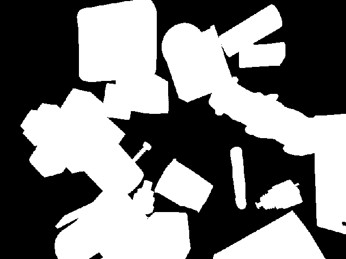}%
    \label{fig4_f}}
    
\caption{Examples of (a) RGB images, (b) ground truth (GT) masks, and (c) perturbed masks used during training. (d-f) the instance representations utilized in QuBER.}
\label{fig4}
\end{figure}

\subsection{Network Architecture}

QuBER comprises three main components (Fig. \ref{fig3}): an IS feature extractor $\mathcal{F}$, an error estimator $\mathcal{E}$, and an error-informed refiner $\mathcal{H}$. We introduce our error-informed refinement scheme with our Error Guidance Fusion (EGF) module, which explicitly incorporates estimated errors into the refinement process. While we implement our approach built on top of Panoptic-DeepLab \cite{cheng2020panoptic} due to its simple lightweight design, the core principles of our error-informed refinement are not inherently tied to this specific architecture.

\textbf{IS Representation.} We represent the initial segmentation using three maps: 1) a center map (Fig. \ref{fig4_d}) indicating the probability of each pixel being an instance center, 2) an offset map (Fig. \ref{fig4_e}) with $x$ and $y$ directions from each pixel to its closest instance center, and 3) a binary foreground (FG) mask (Fig. \ref{fig4_f}). These maps represent both spatial and relational information of the initial segmentation in a fixed-shape format, regardless of the number of instances.

\textbf{IS Feature Extractor.} We employ two parallel ResNet-50 backbones \cite{he2016deep} as RGB-IS and Depth-IS encoders, processing the concatenation of IS with RGB and depth, respectively. The RGB-D-IS fusion decoder combines the RGB-IS and Depth-IS features at residual blocks $C2$, $C3$, and $C5$, using $1\times1$ and $3\times3$ convolutions (only $1\times1$ at $C5$ for efficiency). An atrous spatial pyramid pooling \cite{chen2017deeplab} then extracts multi-scale contextual information. Subsequently, the DeepLab decoder \cite{chen2017deeplab} produces IS features ($w\times h \times c$, where $c$ is the channel dimension), encoding multi-scale texture and geometry information conditioned on IS.

\textbf{Error Estimator.} We implement a lightweight design for the error estimator to predict quadruple boundary errors in the IS. A single $5\times 5$ depthwise separable (DS) convolution \cite{howard2017mobilenets} first extracts error features ($w\times h \times c$). These features are then processed through $1\times 1$ convolutions to predict quadruple boundary errors, guiding the subsequent error-informed refiner to focus on crucial error-containing regions.

\textbf{Error-Informed Refiner:} This component consists of separate branches for predicting foreground, center, and offset maps. We introduce a novel Error Guidance Fusion (EGF) module into each branch to integrate the predicted error estimates with the IS features. The EGF modules take error maps, error features, and IS features as inputs, combining them using a $1\times1$ convolution to reduce channels from $2c+4$ to $c$. This is followed by feature extraction using three $3\times3$ convolutions. This dense fusion of estimated quadruple boundary errors enables targeted and effective adjustments. Each branch then predicts its respective map using $5\times 5$ depthwise separable and $1\times 1$ convolutions. Following \cite{cheng2020panoptic}, post-processing groups foreground pixels with their nearest center to form the final, refined instance masks.

\subsection{Implementation Details}

\textbf{Mask Perturbation.} To enhance the generalization of our model across arbitrary UOIS models, we avoid using initial segmentations from existing models during training. Instead, we apply diverse perturbations to ground truth masks, simulating both fine-grained and instance-level segmentation errors. For fine-grained errors, we employ random contour subsampling, dilation, and erosion operations \cite{cheng2020cascadepsp}. Instance-level errors are simulated through random mask removal and splitting of neighboring masks \cite{xie2022rice}. We also randomly add false positive instances using graph-based segmentation \cite{felzenszwalb2004efficient}. This approach generates a wide distribution of potential segmentation errors, enabling more robust training. Fig. \ref{fig4_c} illustrates an example of our perturbed masks, demonstrating the diversity and realism of the simulated errors.

\textbf{Training.} QuBER, including its error estimator and error-informed refiner, is trained end-to-end. We employ a combination of loss comprising: standard dice loss for error estimation ($L_{err}$), cross-entropy loss for foreground segmentation ($L_{fg}$), MSE loss for center ($L_{ctr}$) \cite{tompson2014joint}, and L1 loss for offset ($L_{off}$) \cite{papandreou2018personlab}. The total loss $L$ is computed as follows:
\begin{equation}
L = \lambda_{err}L_{err} + \lambda_{fg}L_{fg} + \lambda_{ctr}L_{ctr} + \lambda_{off}L_{off}
\end{equation}
We set $\lambda_{err}=1$, $\lambda_{fg}=1$, $\lambda_{ctr}=200$, and $\lambda_{off}=0.01$.

Following standard UOIS protocols \cite{back2022unseen, xiang2021learning, xie2022rice, xie2021unseen}, we train on synthetic data and evaluate on real images without fine-tuning. We use the UOAIS-SIM dataset \cite{back2022unseen}, comprising 50k photorealistic and 100k non-photorealistic synthetic images (Fig. \ref{fig4_a}). Training runs for 90k iterations using Adam optimizer \cite{kingma2014adam} with a learning rate of 0.000125 and a batch size of 8. Color \cite{liu2016ssd} and depth \cite{zakharov2018keep} augmentations are applied for sim-to-real transfer. We use a 640$\times$480 resolution for both training and testing. Training takes approximately 21 hours on two RTX 3090 GPUs. Following \cite{back2022unseen}, we use a lightweight foreground segmentation model \cite{shi2022lmffnet} trained on TOD \cite{xie2020best} to filter background instances (overlap ratio 0.3). Masks smaller than 500 pixels are removed \cite{xie2020best, xie2021unseen}. We use a ResNet-50 backbone pre-trained on ImageNet \cite{deng2009imagenet} and set the center threshold to 0.3 during post-processing.

\section{EXPERIMENTS}
\label{Experiments}

\begin{table*}[t!]
\smallskip
\smallskip
\smallskip
\caption{Performance comparison of QuBER and state-of-the-art methods on OCID using various initial segmentation}
\Large
\centering
\resizebox{0.95\textwidth}{!}{%

\begin{tabular}{|l|r|r|ccc|ccc|ccc|ccc|ccc|}
\hline
\multirow{2}{*}{Method} & \multicolumn{1}{c|}{\multirow{2}{*}{\begin{tabular}[c]{@{}c@{}}Time\\ (ms)\end{tabular}}} & \multicolumn{1}{c|}{\multirow{2}{*}{\begin{tabular}[c]{@{}c@{}}FLOPs\\ (G)\end{tabular}}} & \multicolumn{3}{c|}{Grounded-SAM \cite{ren2024grounded}} & \multicolumn{3}{c|}{UOAIS-Net \cite{back2022unseen}} & \multicolumn{3}{c|}{UOIS-Net-3D \cite{xie2021unseen}} & \multicolumn{3}{c|}{MSMFormer\cite{lu2024mean}} & \multicolumn{3}{c|}{UCN\cite{xiang2021learning}} \\
 & \multicolumn{1}{c|}{} & \multicolumn{1}{c|}{} & $F_O$ & $F_B$ & $F_O^{@.75}$ & $F_O$ & $F_B$ & $F_O^{@.75}$ & $F_O$ & $F_B$ & $F_O^{@.75}$ & $F_O$ & $F_B$ & $F_O^{@.75}$ & $F_O$ & $F_B$ & $F_O^{@.75}$ \\ \hline
Initial Segm. &  &  & 65.1 & 64.9 & 64.5 & 66.7 & 64.4 & 66.5 & 77.4 & 74.2 & 76.3 & 81.9 & 81.3 & 81.6 & 84.1 & 83.0 & 84.9 \\
+ BPR \cite{tang2021look} & 351 & 1722 & 64.6 & 60.0 & 65.2 & 65.2 & 61.5 & 64.8 & 76.9 & 70.9 & 75.6 & 80.1 & 74.2 & 80.8 & 83.2 & 77.5 & 84.5 \\
+ CascadePSP \cite{cheng2020cascadepsp} & 1735 & 40242 & 64.9 & 63.3 & 63.9 & 66.6 & 64.9 & 66.5 & 78.0 & 76.0 & 76.8 & 81.4 & 80.6 & 81.9 & 84.1 & 83.4 & 85.5 \\
+ RICE \cite{xie2022rice} & 2349 & 972 & 72.1 & \textbf{70.4} & 72.5 & 69.4 & 66.6 & 70.0 & 82.9 & 79.1 & 84.3 & 84.8 & 83.4 & 85.7 & 86.4 & \textbf{84.8} & 87.8 \\
+ SAM \cite{kirillov2023segment} & 519 & 5487 & 60.2 & 61.5 & 59.4 & 60.2 & 60.4 & 59.8 & 74.5 & 74.4 & 73.4 & 77.3 & 77.9 & 77.7 & 78.7 & 79.4 & 79.6 \\
+ HQ-SAM \cite{ke2024segment} & 436 & 5520 & 62.6 & 64.1 & 61.6 & 65.6 & 65.5 & 65.0 & 78.2 & 77.2 & 77.0 & 81.2 & 81.5 & 81.6 & 84.1 & 84.7 & 85.1 \\
+ HQ-SAM$\dagger$ \cite{ke2024segment} & 426 & 5519 & 63.9 & 65.1 & 63.0 & 65.8 & 65.5 & 65.3 & 78.3 & 77.3 & 77.1 & 81.3 & 81.6 & 81.8 & 84.5 & 85.1 & 85.6 \\
\textbf{+ QuBER (Ours)} & \textbf{91} & \textbf{431} & \textbf{74.1} & 67.4 & \textbf{73.4} & \textbf{77.8} & \textbf{75.1} & \textbf{76.4} & \textbf{84.5} & \textbf{81.9} & \textbf{85.2} & \textbf{86.2} & \textbf{83.8} & \textbf{87.7} & \textbf{86.6} & 84.3 & \textbf{88.4} \\ \hline
\end{tabular}

}
\label{table1}
\end{table*}

\subsection{Comparison with State-of-the-Art Methods}

We conducted experiments to (1) assess QuBER's effectiveness in refining segmentations from various UOIS models, (2) compare its performance with state-of-the-art refinement methods, and (3) evaluate its consistency across datasets and initial segmentation qualities. Comprehensive experiments were conducted using diverse initial segmentation models and refinement methods across multiple datasets.

\textbf{Datasets.} We evaluate our model on three widely used UOIS benchmark datasets of diverse real cluttered scenes: OCID \cite{suchi2019easylabel}, OSD \cite{richtsfeld2012segmentation}, and WISDOM \cite{danielczuk2019segmenting}. The OCID consists of 2,346 indoor images, including both tabletop and floor scenes, with semi-automated labels. It features an average of 7.5 objects per image, with a maximum of 20 objects. The OSD contains 111 images of tabletop scenes with human-annotated ground truths, having an average of 3.3 objects per image and a maximum of 15 objects. The WISDOM comprises 300 test images of bin scenes with human-annotated ground truths, with an average of 3.8 objects per image, and a maximum of 11 objects. Importantly, the test objects in these datasets do not overlap with those in the training set, allowing for the evaluation on unknown objects.

\textbf{Metrics.} For performance evaluation, we employ standard UOIS metrics \cite{xie2022rice}: object size normalized (OSN) precision, recall, and F-measure for both overlap ($P_O, R_O, F_O$) and boundaries ($P_B, R_B, F_B$). These metrics evaluate segmentation performance over masks and boundaries, accounting fairly for objects of all sizes. Additionally, the percentage of objects with overlap F-measure greater than 0.75 ($F_O^{@.75}$), which evaluates instance-level segmentation accuracy.

\textbf{Initial Segmentation Models.} We employed five state-of-the-art UOIS methods for initial segmentation: Grounded-SAM \cite{ren2024grounded}, UOIS-Net-3D \cite{xie2021unseen}, UOAIS-Net \cite{back2022unseen}, MSMFormer \cite{lu2024mean}, and UCN \cite{xiang2021learning} (the latter two include zoom refinement), using their official implementations. Grounded-SAM, built on SAM, segments objects from RGB images using a given vocabulary; we used a fixed query prompt (a rigid object) following \cite{fang2024embodied}. The other models are category-agnostic UOIS models trained on RGB-D images. We excluded SAM \cite{kirillov2023segment} as an initial segmentation model due to severe over-segmentation without manual prompts ($F_{O}^{@.75}=8.5$, OSD).

\textbf{Segmentation Refinement Baselines.} We compared QuBER with the following state-of-the-art models:
\begin{itemize}
\item BPR \cite{tang2021look}: Crops patches on the boundary of IS masks and refines them using an encoder-decoder network.
\item CascadePSP \cite{cheng2020cascadepsp}: Refines individual IS masks toward fine-grained masks in a coarse-to-fine manner. 
\item RICE \cite{xie2022rice}: An instance-level refinement network that samples instance-level perturbations and selects the best segmentation using a graph neural network.
\item SAM \cite{kirillov2023segment}: A promptable zero-shot segmentation model trained on web-scale datasets (1 billion masks).
\item HQ-SAM \cite{ke2024segment}: An enhanced SAM for high-quality segmentation, with additional high-quality token training. 
\item HQ-SAM$\dagger$ \cite{ke2024segment}: A fine-tuned HQ-SAM for UOIS tasks.
\end{itemize}

To ensure a fair comparison, we used official implementations and evaluated each method's best setup for optimal performance. For BPR and CascadePSP, we trained them on UOAIS-SIM with RGB-D inputs. Official pre-trained weights were used for RICE, SAM, and HQ-SAM. For SAM, HQ-SAM, and HQ-SAM$\dagger$, the ViT-H \cite{dosovitskiy2021an} backbone was used with the box and mask prompts from initial segmentation. HQ-SAM$\dagger$ was fine-tuned on UOAIS-SIM using the token learning approach proposed in the original HQ-SAM paper, starting from the official pre-trained SAM weights.

\textbf{Results.} Table \ref{table1} compares the UOIS performance on the OCID (indoor) dataset of QuBER with state-of-the-art segmentation refinement models over various UOIS models. Tables \ref{table2} and \ref{table3} show results on the OSD (tabletop) and WISDOM (bin) datasets. Across these diverse scenarios, QuBER consistently improves various initial segmentation models and achieves superior performance over all refinement methods. In particular, QuBER achieved superior $F_O^{@.75}$ performance, effectively resolving over-segmentation and under-segmentation issues with error-informed refinement.

We evaluated computational efficiency in Table \ref{table1} using UCN \cite{xiang2021learning} as the IS model. We measured average refinement times and FLOPs from model forward to post-processing on an RTX 3090 and an Intel Xeon Gold 6248R. QuBER demonstrated superior efficiency with inference times under 0.1 seconds and the lowest FLOPs, enabling efficient UOIS refinement. While RICE showed more competitive performance than others due to its instance-level error refinement, its inference time was significantly longer than ours. 

\begin{figure*}[h!]
\smallskip
\smallskip
\centering
  \includegraphics[width=\textwidth]{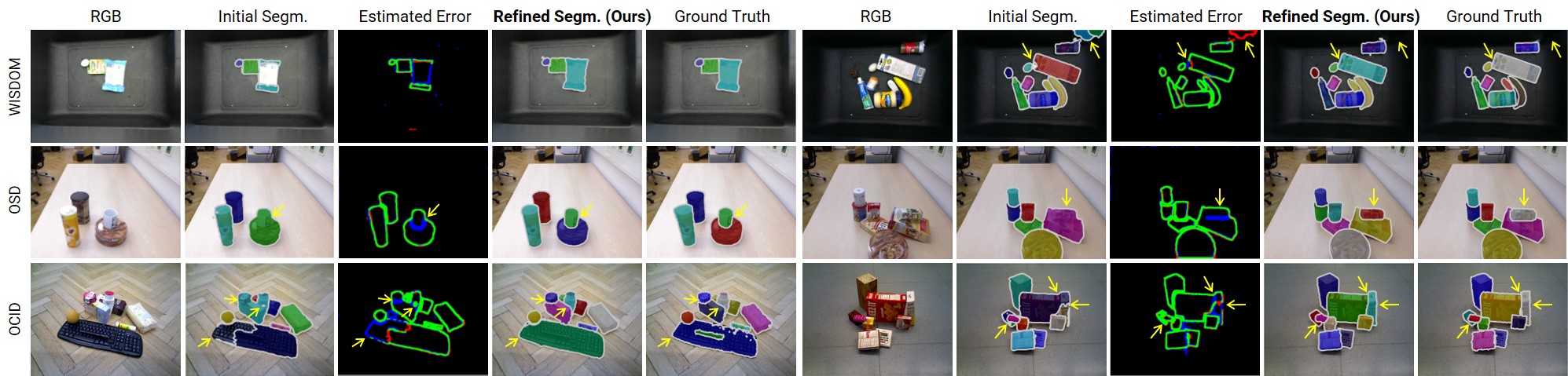}
  \caption{High-quality UOIS results of QuBER on diverse scenes demonstrating accurate object instance segmentation}
  \label{fig5}
\end{figure*}

\begin{figure}[h!]
\smallskip
\smallskip
\centering
  \includegraphics[width=0.45\textwidth]{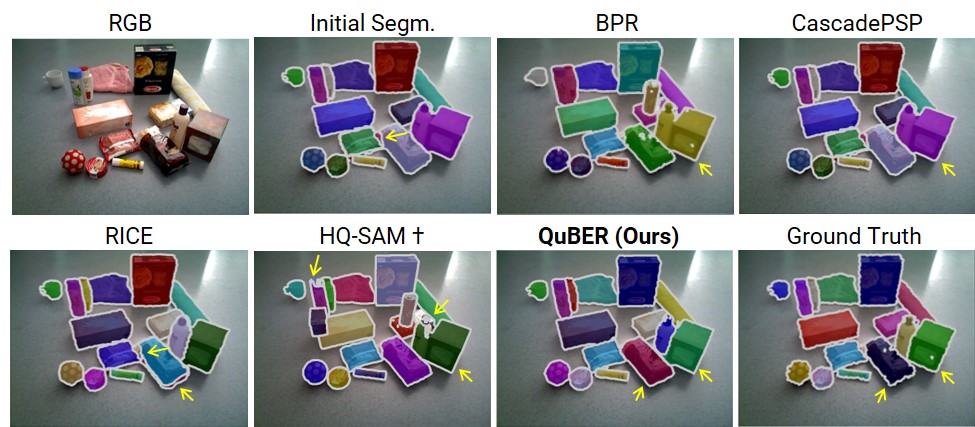}
  \caption{Comparison of QuBER with state-of-the-art methods}
  \label{fig6}
\end{figure}

Fig. \ref{fig5} shows sample qualitative evaluations, and Fig. \ref{fig6} provides a comparison with other state-of-the-art refinement methods using initial segmentation of UCN \cite{xiang2021learning}. The results illustrated in these figures demonstrate that QuBER successfully refines both fine-grained and instance-level errors.

\begin{table}[h!]
\smallskip
\smallskip
\smallskip
\caption{Performance comparison on OSD dataset}
\Large
\centering
\resizebox{0.5\textwidth}{!}{%

\begin{tabular}{|l|ccc|ccc|ccc|}
\hline
\multirow{2}{*}{Method} & \multicolumn{3}{c|}{UOAIS-Net \cite{back2022unseen}} & \multicolumn{3}{c|}{MSMFormer \cite{lu2024mean}} & \multicolumn{3}{c|}{UCN \cite{xiang2021learning}} \\
 & $F_O$ & $F_B$ & $F_O^{@.75}$ & $F_O$ & $F_B$ & $F_O^{@.75}$ & $F_O$ & $F_B$ & $F_O^{@.75}$ \\ \hline
Initial Segm. & 75.7 & 68.6 & 71.5 & 77.0 & 63.4 & 75.7 & 76.4 & 64.7 & 75.5 \\
+ BPR \cite{tang2021look} & 74.7 & 67.6 & 71.3 & 77.6 & 69.3 & 75.9 & 78.7 & 69.9 & 77.1 \\
+ CascadePSP \cite{cheng2020cascadepsp} & 76.3 & 72.3 & 71.6 & 78.8 & 72.5 & 76.2 & 79.3 & 72.8 & 77.6 \\
+ RICE \cite{xie2022rice} & 77.6 & 69.6 & 74.5 & 79.3 & 64.3 & 79.2 & 79.9 & 67.5 & 79.6 \\
+ SAM \cite{kirillov2023segment} & 70.7 & 69.7 & 66.5 & 75.7 & 74.8 & 71.7 & 72.4 & 72.0 & 69.8 \\
+ HQ-SAM \cite{ke2024segment} & 75.5 & 73.8 & 71.3 & 78.1 & 75.6 & 74.5 & 79.0 & 76.8 & 76.5 \\
+ HQ-SAM$\dagger$ \cite{ke2024segment} & 75.6 & 73.2 & 70.5 & 78.9 & 75.7 & 76.2 & 80.2 & 77.5 & 76.9 \\
\textbf{+ QuBER (Ours)} & \textbf{81.4} & \textbf{74.8} & \textbf{78.8} & \textbf{81.4} & \textbf{73.9} & \textbf{79.7} & \textbf{83.8} & \textbf{76.3} & \textbf{83.3} \\ \hline
\end{tabular}

}
\label{table2}
\end{table}

\begin{table}[h!]
\smallskip
\smallskip
\smallskip
\caption{Performance comparison on the WISDOM}
\Large
\centering
\resizebox{0.5\textwidth}{!}{%

\begin{tabular}{|l|ccc|ccc|ccc|}
\hline
\multirow{2}{*}{Method} & \multicolumn{3}{c|}{UOAIS-Net \cite{back2022unseen}} & \multicolumn{3}{c|}{MSMFormer \cite{lu2024mean}} & \multicolumn{3}{c|}{UCN \cite{xiang2021learning}} \\
 & $F_O$ & $F_B$ & $F_O^{@.75}$ & $F_O$ & $F_B$ & $F_O^{@.75}$ & $F_O$ & $F_B$ & $F_O^{@.75}$ \\ \hline
Initial Segm. & 72.8 & 65.7 & 73.3 & 75.3 & 67.5 & 76.8 & 67.8 & 60.0 & 68.0 \\
+ BPR \cite{tang2021look} & 78.4 & 71.0 & 79.8 & 77.0 & 69.4 & 78.9 & 69.9 & 63.2 & 69.8 \\
+ CascadePSP \cite{cheng2020cascadepsp} & 78.4 & \textbf{71.3} & 79.0 & 77.4 & 70.2 & 78.7 & 68.9 & 61.9 & 68.8 \\
+ RICE \cite{xie2022rice} & 78.4 & 70.7 & 79.5 & 78.1 & 69.2 & 80.2 & 75.4 & 66.0 & 75.9 \\
+ SAM \cite{kirillov2023segment} & 67.1 & 63.5 & 67.7 & 70.0 & 66.6 & 70.9 & 63.9 & 60.5 & 63.9 \\
+ HQ-SAM \cite{ke2024segment} & 74.3 & 70.1 & 74.8 & 77.2 & 72.6 & 77.9 & 71.5 & 67.0 & 71.6 \\
+ HQ-SAM$\dagger$ \cite{ke2024segment} & 74.3 & 70.1 & 74.9 & 77.4 & \textbf{72.8} & 78.4 & 71.8 & 67.1 & 72.2 \\
\textbf{+ QuBER (Ours)} & \textbf{78.5} & 70.9 & \textbf{80.8} & \textbf{79.7} & 71.7 & \textbf{81.9} & \textbf{76.4} & \textbf{68.5} & \textbf{77.5} \\ \hline
\end{tabular}

}
\label{table3}
\end{table}

\begin{table}[h!]
\caption{Ablation of error-informed refinement on OCID (PDL: Panoptic-DeepLab)}
\centering
\resizebox{\columnwidth}{!}{%
\begin{tabular}{|l|ccc|lll|}
\hline
\multirow{2}{*}{Method} & segm. & error & \multirow{2}{*}{EGF} & \multirow{2}{*}{$F_O$} & \multirow{2}{*}{$F_B$} & \multirow{2}{*}{$F_O^{@.75}$} \\
 & refine. & estim. &  &  &  &  \\ \hline
UCN \cite{xiang2021learning} & \textcolor{Red}{\xmark} & \textcolor{Red}{\xmark} & \textcolor{Red}{\xmark} & 84.1 & 83.0 & 84.9 \\
PDL \cite{cheng2020panoptic} & \textcolor{Red}{\xmark} & \textcolor{Red}{\xmark} & \textcolor{Red}{\xmark} & 80.4 \scriptsize(\textcolor{blue}{ -3.7}) & 71.3 \scriptsize(\textcolor{blue}{ -11.7}) & 76.4 \scriptsize(\textcolor{blue}{ -8.5}) \\
UCN \cite{xiang2021learning} + PDL \cite{cheng2020panoptic} & \textcolor{Green}{\cmark} & \textcolor{Red}{\xmark} & \textcolor{Red}{\xmark} & 82.2 \scriptsize(\textcolor{blue}{ -1.9}) & 77.2 \scriptsize(\textcolor{blue}{ -5.8}) & 81.0 \scriptsize(\textcolor{blue}{ -3.9}) \\
UCN \cite{xiang2021learning} + \textbf{QuBER} & \textcolor{Green}{\cmark} & \textcolor{Green}{\cmark} & \textcolor{Red}{\xmark} & 84.7 \scriptsize(\textcolor{red}{+0.6}) & 81.3 \scriptsize(\textcolor{blue}{ -1.7}) & 85.3 \scriptsize(\textcolor{red}{+0.4}) \\
UCN \cite{xiang2021learning} + \textbf{QuBER} & \textcolor{Green}{\cmark} & \textcolor{Green}{\cmark} & \textcolor{Green}{\cmark} & \textbf{86.1 \scriptsize(\textcolor{red}{+2.0})} & \textbf{83.7 \scriptsize(\textcolor{red}{+0.7})} & \textbf{87.6 \scriptsize(\textcolor{red}{+2.6})} \\ \hline
\end{tabular}
}
\label{table4}
\end{table}

\subsection{Ablation Studies}
\textbf{Effect of Error-informed Refinement.} We evaluated the impact of our proposed error-informed refinement by comparing QuBER with variants lacking the error estimation and EGF modules (Table \ref{table4}). We used the original Panoptic-DeepLab \cite{cheng2020panoptic} as a baseline, evaluating it as both an IS and a refinement model with RGB-D inputs, similar to QuBER's setup. Results show that Panoptic-DeepLab without error estimation and EGF fails to achieve state-of-the-art performance as both an IS and a refinement model, even underperforming UCN. In contrast, QuBER, which incorporates both error estimation and EGF, significantly outperforms the others, demonstrating the effectiveness of our error-informed refinement scheme for high-quality UOIS.

\textbf{Effect of Quadruple Boundary Error.} To assess the importance of our novel quadruple boundary error, we conducted an ablation study on the OCID dataset using UCN as the IS model (Table \ref{table5}). We compared three error estimation approaches: binary boundary error, mask quadruple error, and our proposed quadruple boundary error (first to third rows, respectively). Results clearly demonstrate that our quadruple boundary error yields the best performance, highlighting its effectiveness in capturing complex segmentation errors and guiding precise refinement.

\begin{table}[h!]
\caption{Ablation of quadruple boundary error on OCID}
\centering
\resizebox{0.9\columnwidth}{!}{%
    \begin{tabular}{|cc|ccc|ccc|c|}
    \hline
    Quadruple & Boundary & \multicolumn{3}{c|}{Overlap} & \multicolumn{3}{c|}{Boundary} &  \\
    Error & Error & $P_N$ & $R_O$ & $F_O$ & $P_B$ & $R_B$ & $F_B$ & $F_O^{@.75}$ \\ \hline
    \textcolor{Red}{\xmark} & \textcolor{Green}{\cmark} & 88.2 & 88.5 & 84.6 & 84.0 & 85.3 & 81.5 & 85.4 \\
    \textcolor{Green}{\cmark} & \textcolor{Red}{\xmark} & \textbf{88.9} & 89.3 & 85.8 & 85.7 & 86.4 & 83.1 & 87.3 \\
    \textcolor{Green}{\cmark} & \textcolor{Green}{\cmark} & \textbf{88.9} & \textbf{89.8} & \textbf{86.1} & \textbf{86.1} & \textbf{87.1} & \textbf{83.7} & \textbf{87.6} \\ \hline
    \end{tabular}
}
\label{table5}
\end{table}

\subsection{Robot Experiments: Target Object Grasping}

We evaluated the effect of QuBER on the performance of UOIS in a practical robotic task—segmenting and grasping unknown target objects from cluttered bins—a scenario common in warehouse pick-and-place \cite{mitash2023armbench}. By integrating QuBER into state-of-the-art UOIS methods (UOAIS-Net \cite{back2022unseen} and UCN \cite{xia2020synthesize}), we aimed to demonstrate its ability to improve segmentation quality and grasping success rates.

\begin{figure}[h!]
    \centering
     \includegraphics[width=\columnwidth]{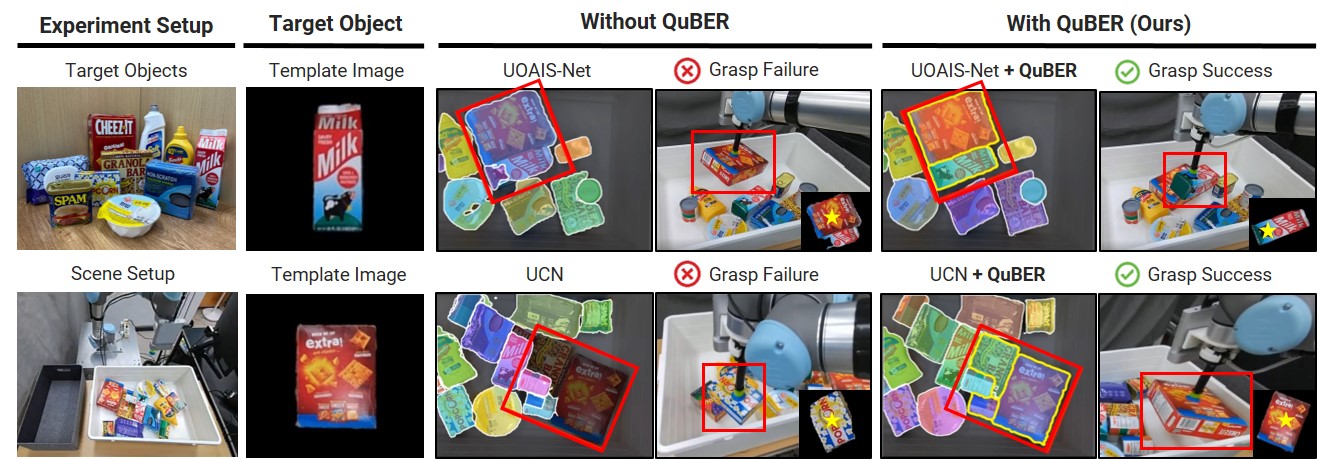}
    \caption{Target object grasping setup and results. QuBER refines initial UOIS results, leading to the successful grasping of the target object. (Yellow star: grasping point from segmentation. Videos are available in the supplementary.)}
    \label{fig7}
\end{figure}

\textbf{Setup.} We used a UR5 robotic arm with an Azure Kinect camera mounted for hand-eye coordination and a suction gripper. In each trial, up to 20 objects were randomly placed in a bin (Fig. \ref{fig7}). To simulate real-world identification constraints, we used ten template images per target object for matching. The pipeline involved three steps: 1) performing UOIS on RGB-D images using either UOAIS-Net or UCN, both with and without QuBER refinement; 2) matching the segmented objects to the templates using cosine distance between pre-trained DINOv2 features \cite{oquab2023dinov2}; and 3) executing a suction grasp on the most planar regions using RANSAC plane fitting \cite{gouda2022dopose}. We conducted 100 trials per method, attempting ten grasps for each of ten distinct target objects. To ensure a fair comparison, object placements remained consistent across trials with and without QuBER refinement.

\textbf{Results.} Table \ref{table8} presents our findings, evaluated using two metrics: segmentation success rate (accurate segmentation and matching without over- or under-segmentation) and grasp success rate (successful target object removal from the bin). QuBER consistently improved both metrics across all experiments. For UCN, QuBER improved overall performance by detecting missing instances. For UOAIS-Net, QuBER effectively resolved over- and under-segmentation issues, enhancing object matching and grasping performance.

\begin{table}[h!]
\caption{Target object grasping performances}
\centering
\resizebox{0.7\columnwidth}{!}{%
\begin{tabular}{l|ll}
\hline
\multirow{2}{*}{Method} & segmentation & grasp \\
 & success rate & success rate \\ \hline
UCN \cite{xie2021unseen} & 65\% & 59\% \\
+ QuBER & \textbf{82\%} (\textcolor{red}{+17\%}) & \textbf{75\%} (\textcolor{red}{+16\%}) \\ \hline
UOAIS-Net \cite{back2022unseen} &  80\% & 70\% \\
+ QuBER & \textbf{86\%} (\textcolor{red}{+6\%}) & \textbf{76\%} (\textcolor{red}{+6\%}) \\ \hline
\end{tabular}
}
\label{table8}
\end{table}


\section{CONCLUSION AND FUTURE WORK}

We introduced QuBER, an error-informed refinement method for high-quality UOIS. With quadruple boundary error estimation and EGF module, QuBER achieved state-of-the-art segmentation and enhanced robotic grasping performance. However, estimated errors may still contain inaccuracies, especially in cases of severe occlusion and out-of-distribution. Future work will focus on scaling up datasets and incorporating continual learning to improve robustness.

\section*{ACKNOWLEDGMENT}
\begin{spacing}{0.5}
{\scriptsize This work was supported by the Technology Innovation Program (Project Number: 
 00442029) funded by the Ministry of Trade Industry \& Energy (MOTIE, Korea), and Korea Institute for Advancement of Technology (KIAT) grant funded by the Korea Government (MOTIE) (Project Number: 20008613). This work was partially supported by Artificial intelligence industrial convergence cluster development project funded by the Ministry of Science and ICT (MSIT, Korea) \& Gwangju Metropolitan City.}
\end{spacing}

\FloatBarrier
\bibliographystyle{IEEEtran}
\bibliography{references.bib}

\end{document}